\documentclass[conference]{IEEEtran}
\IEEEoverridecommandlockouts
\usepackage{subfig}
\usepackage{multirow}
\usepackage{balance}
\usepackage{cite}
\usepackage{amsmath,amssymb,amsfonts}
\usepackage{algorithmic}
\usepackage{graphicx}
\usepackage{textcomp}
\usepackage{xcolor}
\def\BibTeX{{\rm B\kern-.05em{\sc i\kern-.025em b}\kern-.08em
    T\kern-.1667em\lower.7ex\hbox{E}\kern-.125emX}}
\begin{document}

\title{FMCW Radar Sensing for Indoor Drones Using Learned Representations
\thanks{This research received funding from the Flemish Government (AI Research Program). Ozan \c{C}atal was funded by a Ph.D. grant of the Flanders Research Foundation (FWO).}
}

\author{
\IEEEauthorblockN{Ali Safa$^{1,3}$, Tim Verbelen$^{2,3}$, Ozan \c{C}atal$^{2,3}$, Toon Van de Maele$^{2}$, 
\\Matthias Hartmann$^{3}$, Bart Dhoedt$^{2,3}$, André Bourdoux$^{3}$\\
\textit{$^{1}$ESAT, KU Leuven, $^{2}$IDLab, Ghent University, $^{3}$imec, 3001, Leuven, Belgium}}
\{Ali.Safa, Tim.Verbelen\}@imec.be
}



\maketitle
\thispagestyle{empty}
\pagestyle{empty}

\begin{abstract}

Frequency-modulated continuous-wave (FMCW) radar is a promising sensor technology for indoor drones as it provides range, angular as well as Doppler-velocity information about obstacles in the environment. Recently, deep learning approaches have been proposed for processing FMCW data, outperforming traditional detection techniques on range-Doppler or range-azimuth maps. However, these techniques come at a cost; for each novel task a deep neural network architecture has to be trained on high-dimensional input data, stressing both data bandwidth and processing budget. In this paper, we investigate unsupervised learning techniques that generate low-dimensional representations from FMCW radar data, and evaluate to what extent these representations can be reused for multiple downstream tasks. To this end, we introduce a novel dataset of raw radar ADC data recorded from a radar mounted on a flying drone platform in an indoor environment, together with ground truth detection targets. We show with real radar data that, utilizing our learned representations, we match the performance of conventional radar processing techniques and that our model can be trained on different input modalities such as raw ADC samples of only two consecutively transmitted chirps.
\end{abstract}

\section*{Supplementary Material}
We release the dataset used in this work in the link below$^1$.
\section{Introduction}
Indoor flying with drones is significantly more difficult than flying outdoors, due to the lack of exact positioning information such as GPS, and the more stringent requirements on obstacle detection and avoidance~\cite{Indoor}. Building a coherent view of the world from sensory observations is therefore an important challenge for autonomous indoor drones. An often overlooked sensor, which is useful for this purpose, is the frequency-modulated continuous-wave (FMCW) radar. The ability to get robust range, velocity and angle estimates from a single radar frame provides excellent additional information to the traditionally used accelerometer, gyro and camera sensors. Conventional radar processing approaches process raw radar ADC values through Fourier processing into range, velocity and angular information~\cite{richards2005fundamentals}. However, the resulting data format is still high dimensional and thus also requires high bandwidth connections between the radar sensor and the downstream processing unit. Often, this Fourier processing is followed by Constant False Alarm Rate (CFAR) algorithms that result in a list of detected targets, their distance, velocity and angle of arrival~\cite{Safa21}.

Recently, deep learning techniques have proven to be well suited for various sensor processing tasks. This is also the case for radar, where their feature learning capabilities often make them better suited than classical algorithms, outperforming techniques based on handcrafted feature extraction~\cite{Wheeler2017}. Such deep neural networks (DNNs) typically operate on the Fourier processed data, such as range-azimuth-Doppler maps~\cite{Major2019} or micro-Doppler maps~\cite{Prahbat21}. The weakness of these approaches, however, is that retraining of the DNN on high-dimensional preprocessed radar maps is needed for each task at hand.

In this paper, we investigate whether \textit{unsupervised} deep learning approaches can be leveraged to yield a low-dimensional radar representation that can be reused for a wide range of downstream tasks. In addition, we also research to what extent such representations can be learned from raw ADC data of \textit{only two chirps}, instead of a complete chirp frame typically used in radar processing~\cite{Major2019,Prahbat21}, potentially saving energy usage and data bandwidth. To this end, we focus on an indoor drone flight scenario, and introduce a novel large-scale dataset of raw FMCW radar samples as well as ground truth information for a number of downstream target detection tasks. Our results show that low-dimensional, unsupervised learned representations from range-Doppler and range-angular maps can successfully embed range, velocity and angular information of targets in the environment. Also representations trained on data from only two chirps contain relevant range, velocity and angular information.

To summarize, we make the following contributions:
\begin{itemize}
    \item We present a novel, large scale dataset of RGBD and raw FMCW radar data captured on an indoor flying drone~\footnote{https://thesmartrobot.github.io/datasets}.
    \item We propose a generic unsupervised learning method for acquiring low-dimensional representations of radar data given different input formats.
    \item We evaluate the effectiveness of these representations on a set of downstream tasks, and compare our results against traditional radar processing techniques.
    \item We show that our method can successfully perform with even raw ADC samples of only two consecutive chirps.
\end{itemize}

This paper is organized as follows. Related work is covered in Section \ref{related}. Our dataset acquisition is explained in Section \ref{dataset}. Our various neural network models are presented in Section \ref{models}. Experimental results are shown in Section \ref{experiments}. Conclusions are provided in Section \ref{conc}.

    

\begin{figure*}[t!]
    \centering
    \subfloat[]{\includegraphics[height=7.5em]{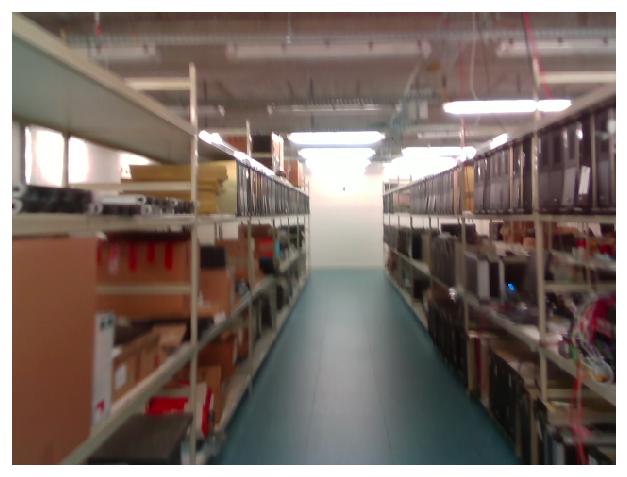}} \hfill
    \subfloat[]{\includegraphics[height=7.5em]{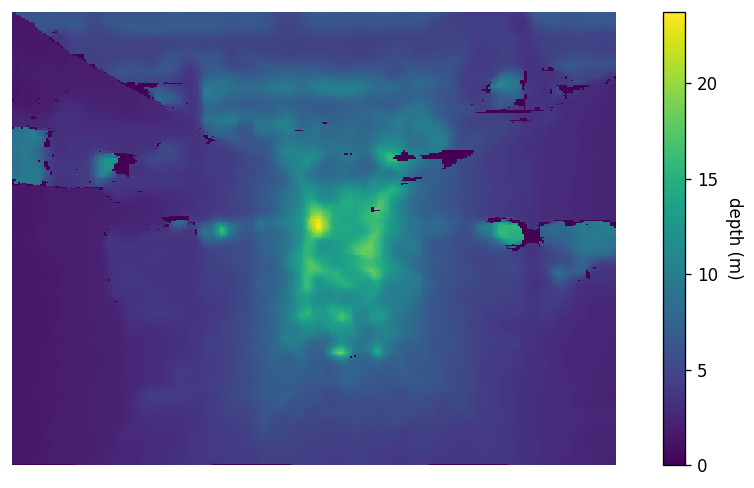}} \hfill
    \subfloat[]{\includegraphics[height=7.5em]{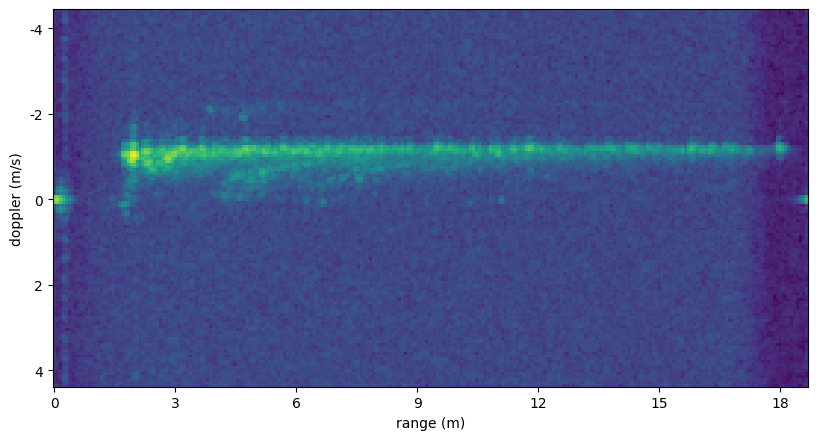}} \hfill
    \subfloat[]{\includegraphics[height=7.5em]{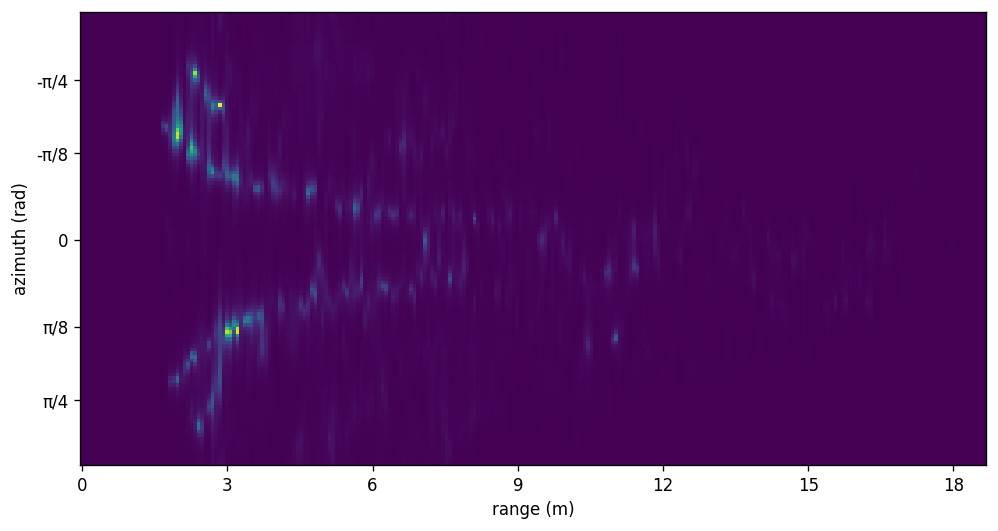}} \hfill
    \\
    \subfloat[camera]{\includegraphics[height=7.5em]{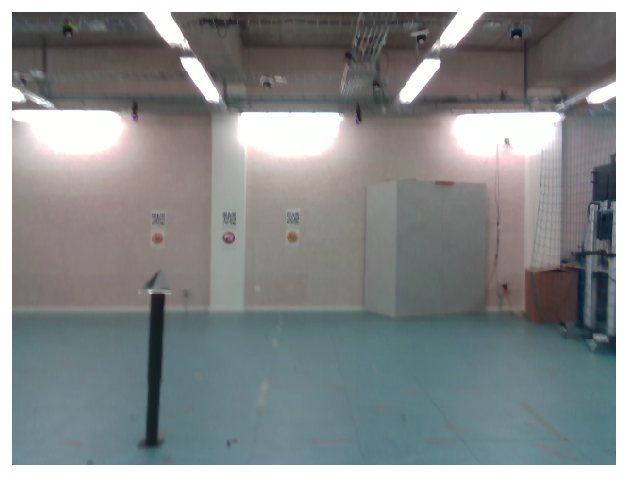}} \hfill
    \subfloat[depth]{\includegraphics[height=7.5em]{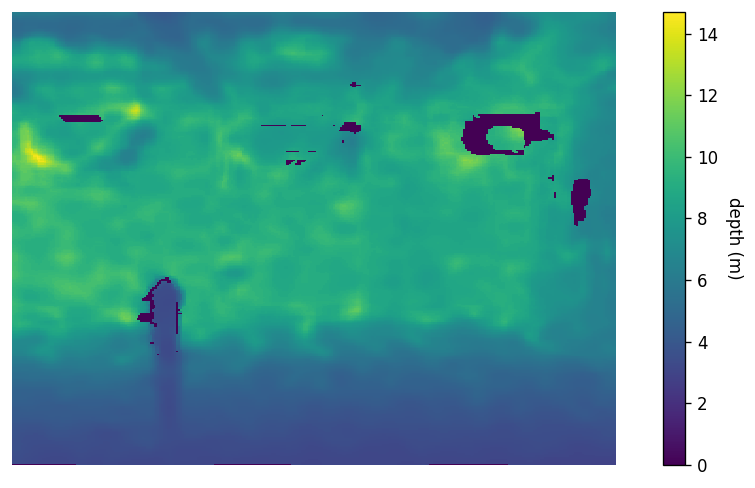}} \hfill
    \subfloat[range-Doppler]{\includegraphics[height=7.5em]{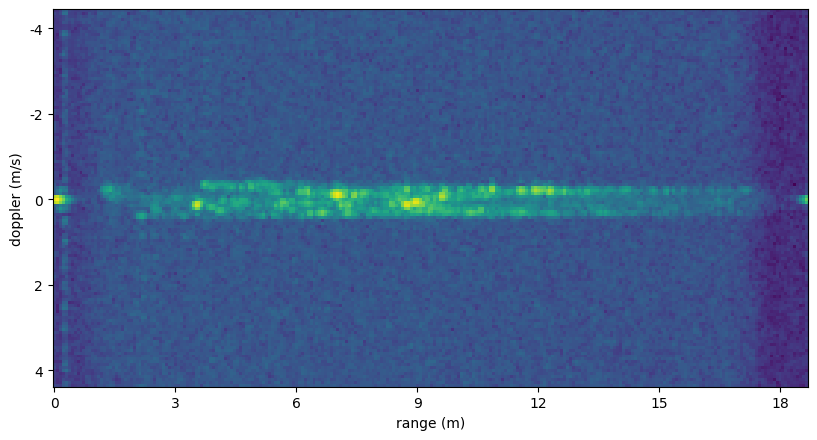}} \hfill
    \subfloat[range-azimuth]{\includegraphics[height=7.5em]{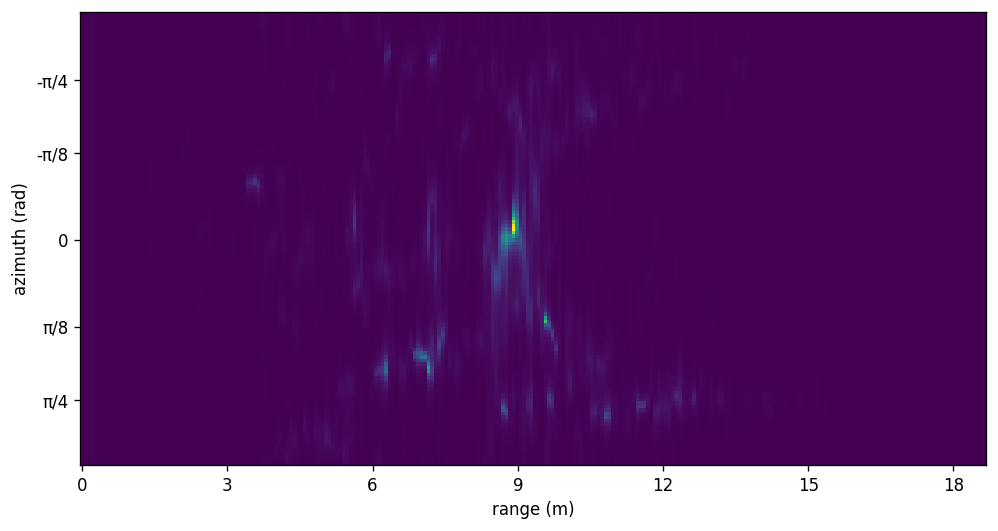}} \hfill
    \caption{Example of camera, depth, range-Doppler and range-azimuth views from scenario 1 (a,b,c,d), flying through the aisles, and scenario 3 (e,f,g,h), hovering in front of a corner reflector.}
    \label{fig:data}
\end{figure*}

\section{Related work}
\label{related}
FMCW radar has been extensively used for range, velocity and angle-of-arrival detection~\cite{richards2005fundamentals}, especially in the context of automotive applications. Recently, there has been increasing interest in utilizing deep neural networks to improve radar processing, i.e. focusing on radar antenna design, radar signal recognition, automatic target recognition based on high range resolution profiles, and clutter recognition and suppression~\cite{Geng2021}.

Most related work operates in the application area of autonomous driving. For example Major et al.~\cite{Major2019} used a convolutional neural network that processes range-azimuth-Doppler tensors for vehicle detection. Patel et al.~\cite{Patel2019} similarly utilize a convolutional neural network for object classification in a supervised manner, while Wheeler et al.~\cite{Wheeler2017} proposed a VAE architecture fine-tuned to range-azimuth maps of autonomous driving data to learn an accurate radar sensor model.
In the context of indoor flying drones, radars have been used as a sensor for obstacle detection~\cite{Safa21} and odometry estimation~\cite{Mostafa2018}. Deep learning techniques have been further proposed for localization and activity classification of drones using micro-Doppler signatures~\cite{Prahbat21}.

The main drawback of current deep learning approaches for radar is that they operate on the high-dimensional radar maps, and have to be retrained for each novel task. In contrast, we propose to learn generic low-dimensional radar representations that can be reused for a number of downstream tasks. 

\section{Dataset}
\label{dataset}
\begin{figure}[b!]
        \centering
        \includegraphics[width=0.45\textwidth]{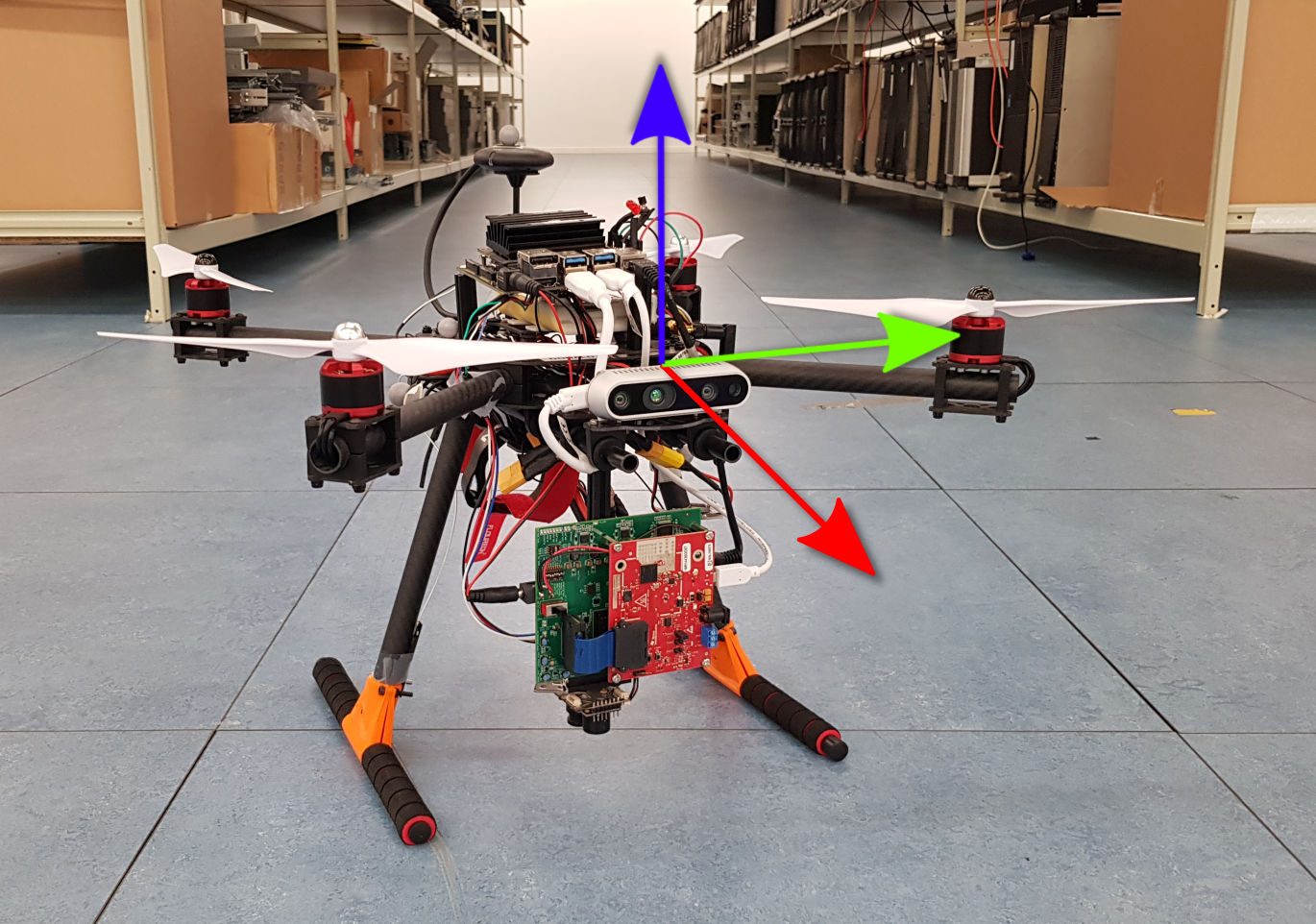}
        \caption{We use an NXP drone equipped with TI 79GHz mmWave radar and Intel Realsense D435 RGBD camera. The drone reference frame is tracked by a Qualisys MOCAP system when in view of the cameras.}
        \label{fig:drone}
\end{figure}

We collected a novel indoor drone flying dataset using an NXP hovergames drone in a warehouse lab setting. This dataset is the first of its kind since it contains both RGBD camera data as well as FMCW radar. In addition, we log the raw ADC radar samples, in contrast to the more common  post processed radar maps. Our drone is shown in Fig.~\ref{fig:drone}, and carries both a TI IWR 1443 mmWave radar and an Intel Realsense D435 camera. The specific sensor configuration parameters used are given in Table~\ref{tab:params}. In addition to RGBD and FMCW data, we also store the internal Extended Kalman Filter (EKF) state of the PX4 flight controller, as well as the actuator commands (yaw-pitch-roll-thrust) sent by the radio remote control. All data is recorded on a Jetson Nano on-board computer, and is synchronized to the radar frame rate of 5 FPS. We also record a 6-DOF pose using a Qualisys marker-based motion capture system (MOCAP), providing a ground truth signal for a subset of the downstream regression tasks defined in Section~\ref{sec:tasks}.

\begin{table}[t!]
    \centering
    \begin{tabular}{| l  c | }
       \hline
       \multicolumn{2}{|c|}{\textbf{TI IWR 1443 mmWave radar}} \\
       \hline
       number of chirps & 128 \\
       number of transmit antennas & 2 \\
       number of receive antennas & 4 \\
       number of samples per chirp & 256 \\
       start frequency & 77GHz \\
       frequency slope & 50e12 \\
       sample rate & 6.24e6 \\
       ADC start time & 11 $\mu$s \\
       ramp end time & 68 $\mu$s \\
       idle time & 40 $\mu$s \\
       \hline
       \multicolumn{2}{|c|}{\textbf{Intel Realsense D435 camera}} \\
       \hline
       color resolution (W x H) & 640 $\times$ 480 \\
       depth resolution (W x H) & 640 $\times$ 480 \\
       \hline
    \end{tabular}
    \caption{Sensors configuration.}
    \label{tab:params}
\end{table}

\begin{figure}[t!]
    \centering
    \subfloat[]{\includegraphics[height=18em]{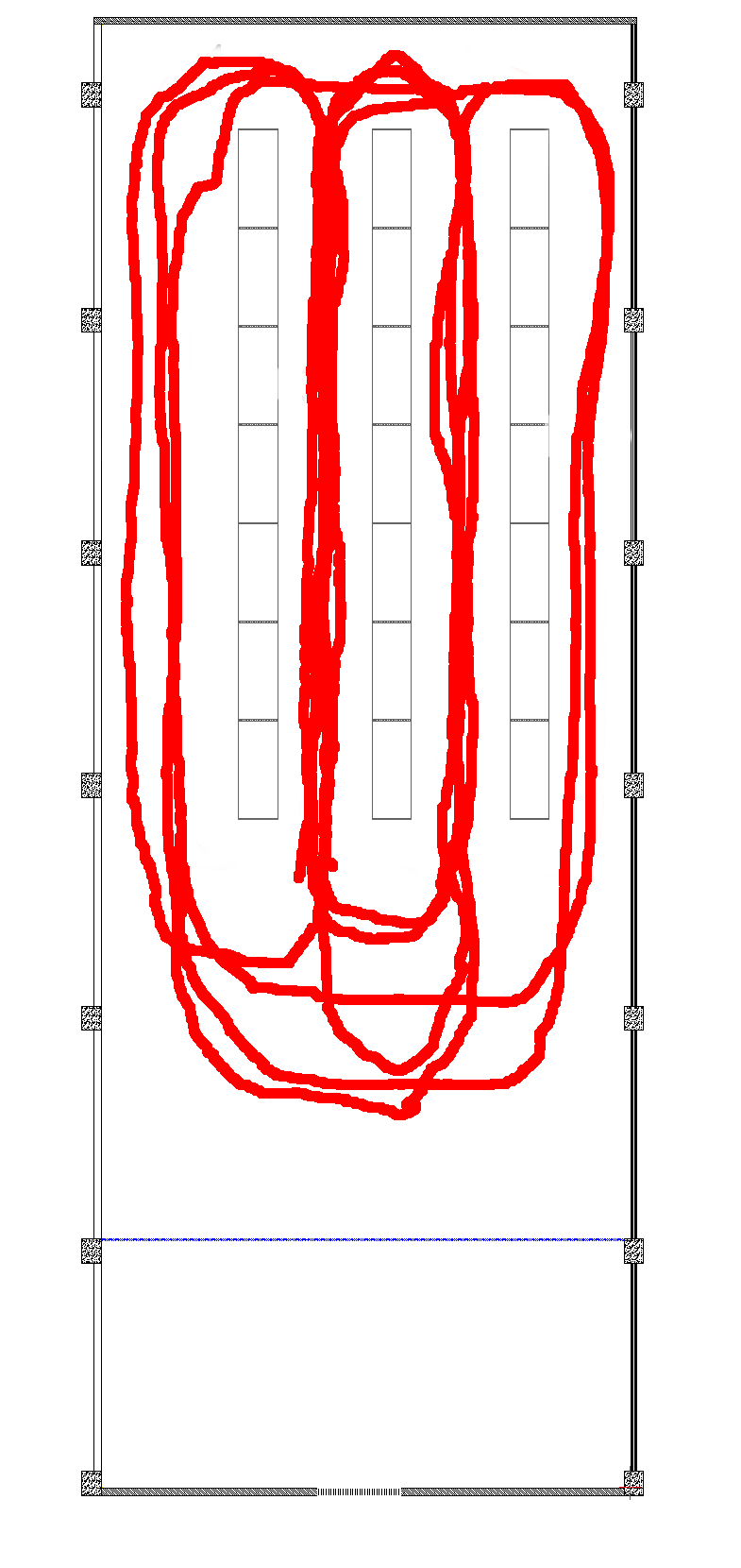}}
    \subfloat[]{\includegraphics[height=18em]{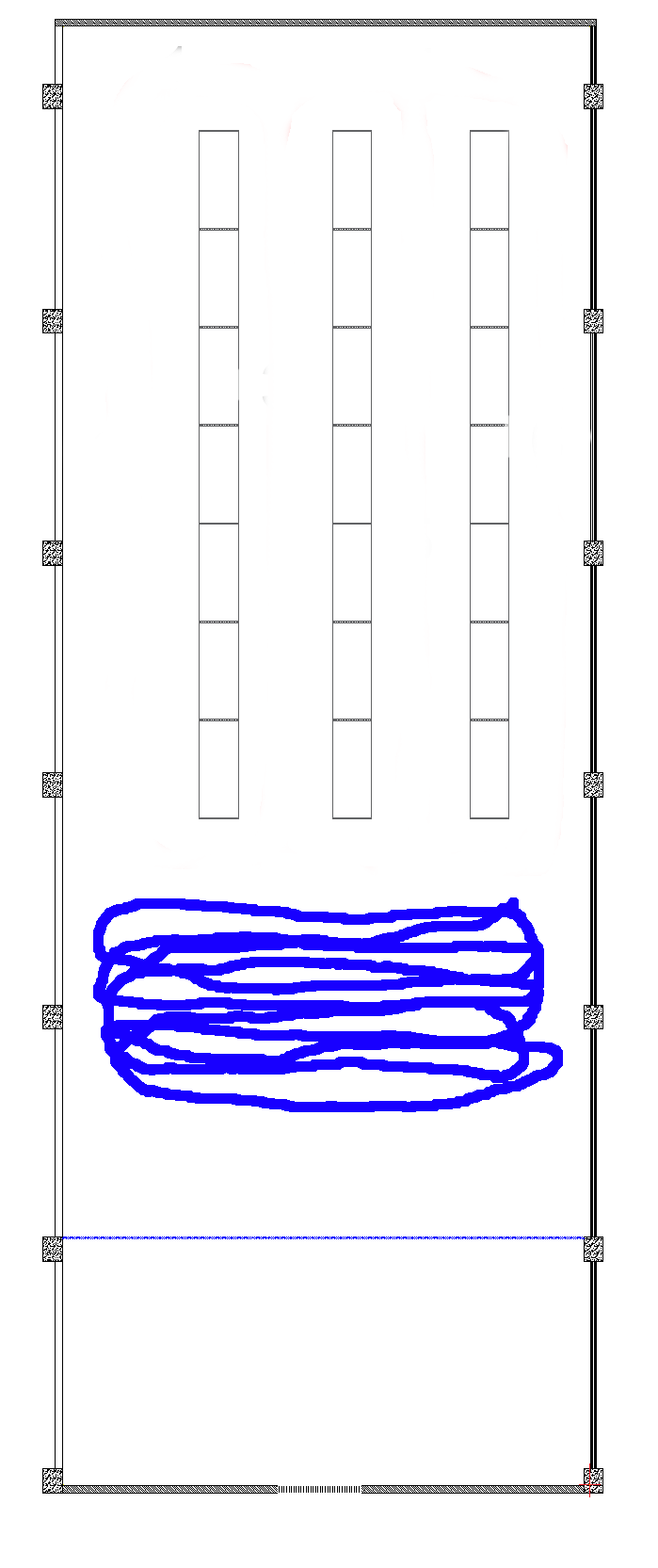}}
    \subfloat[]{\includegraphics[height=18em]{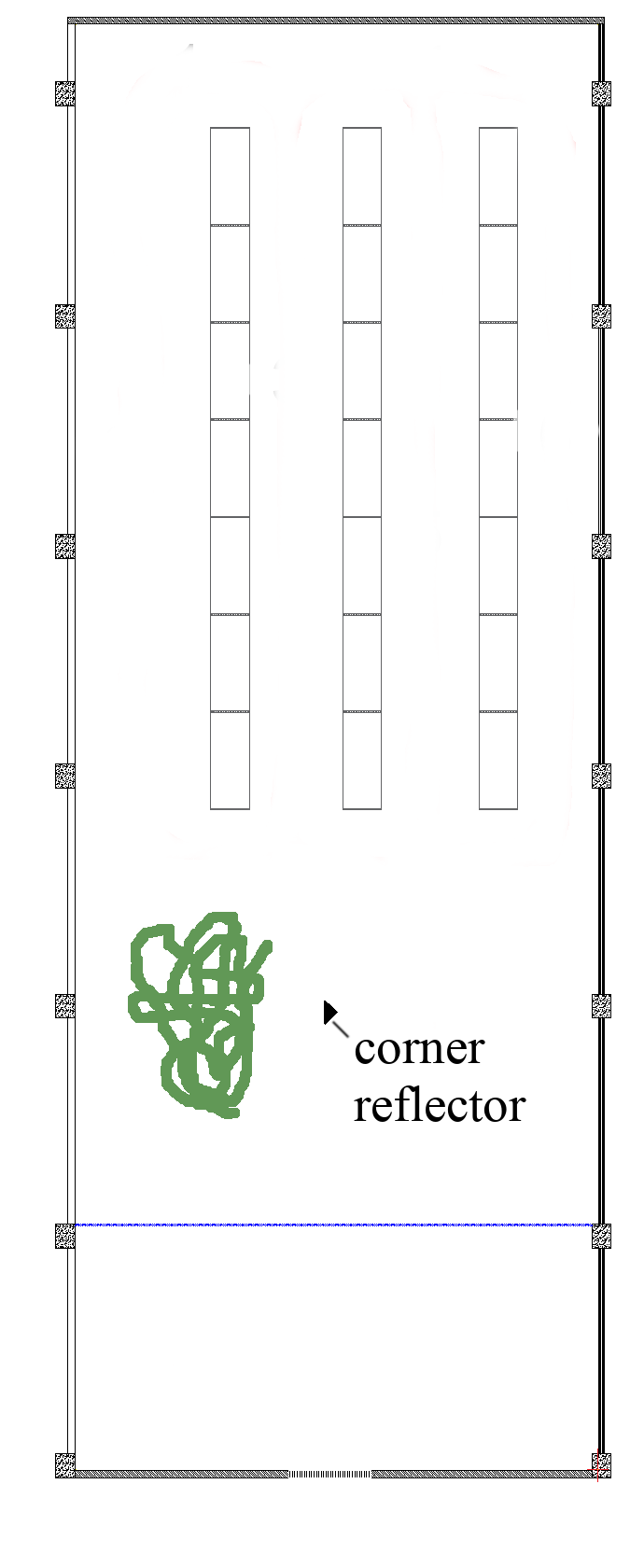}}   
    \caption{We record data in three scenarios: (a) scenario 1: flying trajectories through the aisles, (b) scenario 2: flying from one wall to another, and (c) scenario 3: hovering in front of a corner reflector.}
    \label{fig:scenarios}
\end{figure}

Data is collected in three recording scenarios in which the drone is manually flown through the lab, as shown in Fig.~\ref{fig:scenarios}). In the first scenario, the drone flies throughout the lab, in between narrow aisles mimicking a warehouse layout. In the second scenario, the drone flies from wall to wall in the open space, tracked by the MOCAP system. In the third scenario, a corner reflector, i.e. a aluminium foil pyramidal reflector (also depicted on Fig.\ref{fig:data}e), is placed at the center of the open space (i.e. at the origin of the MOCAP reference frame), and the drone hovers in front of the corner reflector. For scenario 1 to 3, we collected 11274, 5005 and 2798 frames, respectively, totaling 38GB of data. 

In addition to the raw ADC samples, we also include post-processed range-Doppler and range-azimuth maps in the dataset. The range-Doppler maps are generated by 2D Fourier transforming the raw ADC samples over the fast- and slow-time, and then averaging the amplitude over all antennas. For the range-azimuth map, we use Capon beamforming~\cite{Capon}, with 128 angle bins, resulting in range-azimuth maps of identical size as the range-Doppler maps. This allows us to re-use the same neural network architecture for both modalities as described later. Fig.~\ref{fig:data} shows examples of the collected dataset, with the camera and depth data, as well as corresponding range-Doppler and range-azimuth maps of the radar data.

\section{Models}
\label{models}
\begin{figure}[b!]
    \centering
    \subfloat[RD VAE]{\includegraphics[width=0.4\textwidth]{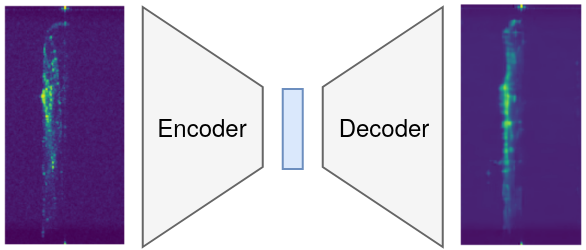}}\\
    \subfloat[RA VAE]{\includegraphics[width=0.4\textwidth]{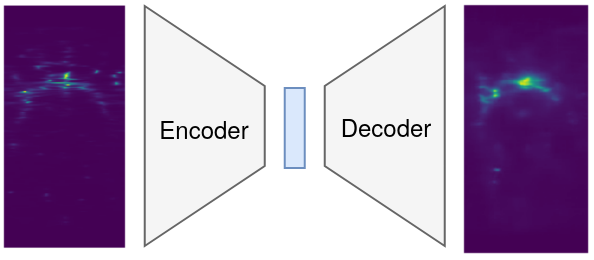}}\\
    \subfloat[Chirp VAE]{\includegraphics[width=0.3\textwidth]{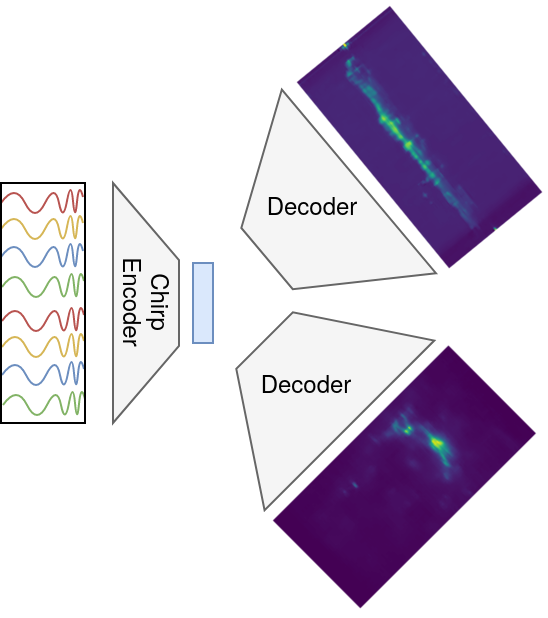}}   
    \caption{The different proposed sensor models. The range-Doppler (RD) VAE (figure a) takes a range-Doppler map as input. The range-azimuth (RA) VAE (figure b) takes a range-azimuth map as input. The chirp VAE (figure c) takes the ADC samples of two consecutively transmitted chirps as input and reconstructs both the range-Doppler and range-azimuth maps corresponding to that observation.}
    \label{fig:model high level}
\end{figure}

As we are interested in generating relevant intermediate representations suitable for a wide array of downstream tasks, we propose three different generative models for radar processing, all based on the well-established variational autoencoder (VAE) framework~\cite{Kingma2014,Rezende2014}. The high-level data-flow is shown in Fig.\ref{fig:model high level}. Similarly to existing related work on object detection in automotive scenarios, we build a VAE model that operates on range-azimuth (RA) and range-Doppler (RD) maps. Finally, we also propose a novel approach to radar processing which leverages a generative modelling approach to generating RA and RD maps from raw ADC samples directly.

\subsection{Variational Autoencoders}
All our proposed architectures are based on the Variational AutoEncoder (VAE) framework for generative modelling~\cite{Kingma2014,Rezende2014}. A VAE amortizes the variational Bayesian inference process using two neural networks: an encoder and decoder network. These are jointly optimized by maximizing the evidence lower bound (ELBO), or equivalently by minimizing the following loss function:\
\begin{equation}
    \mathcal{L} = E_q\big[\log p_\theta(x | z) \big] - D_{KL} \big[q_\phi(z | x) || p(z) \big],
\end{equation}
where $p(z)$ represents a (fixed) prior distribution, $p_\theta(x|z)$ the learned likelihood (decoder) distribution, and $q_\phi(z | x)$ the learned variational posterior (encoder) distribution, with $x$ and $z$ the observed and latent variables respectively. Concretely, the encoder maps high-dimensional sensor inputs to a low dimensional state space, which is parameterized as a multivariate, isotropic Gaussian by outputting distribution means and standard deviations from the encoder model. As a prior distribution, a standard Normal distribution is typically chosen $p(z) = \mathcal{N}(0,1)$.

\subsection{Range-Doppler VAE}
The first model (Fig.~\ref{fig:model high level}a) we investigate uses the range-Doppler maps as input and target output. The input RD-map is generated out of the raw radar ADC samples through range-Doppler Fourier processing~\cite{Safa21}. As likelihood loss the mean squared error (MSE) is used. The model is parameterized using convolutional neural nets using 32 latent dimensions with the architecture described in Table~\ref{tab:RAArchitectures}.

\begin{table}[b!]
    \centering
    \begin{tabular}{l | c | c | c  }
        & Layer & Neurons/Filters & activation function  \\\hline \hline
        \parbox[t]{2mm}{\multirow{7}{*}{\rotatebox[origin=c]{90}{RA/RD Encoder}}}
        &Convolutional & 16  & Leaky ReLU \\
        &Convolutional & 16  & Leaky ReLU \\
        &Convolutional & 32  & Leaky ReLU \\
        &Convolutional & 32  & Leaky ReLU \\
        &Linear        & 1024& Leaky ReLU      \\ 
        &Linear        & 64  & None (mean) and Softplus (stdev) \\
        &&&\\\hline
        
        \parbox[t]{2mm}{\multirow{6}{*}{\rotatebox[origin=c]{90}{Chirp Encoder}}}
        &Convolutional & 512 & Leaky ReLU \\
        &Convolutional & 512 & Leaky ReLU \\
        &Convolutoinal & 32 & Leaky ReLU \\
        &Linear        & 512 & Leaky ReLU \\
        &Linear        & 64  & None (mean) and Softplus (stdev) \\&&&\\\hline
        
        \parbox[t]{2mm}{\multirow{7}{*}{\rotatebox[origin=c]{90}{Decoder}}}
        & Linear &  1024    & Leaky ReLU \\
        & Linear & 4096     & Leaky ReLU \\
        &Convolutional & 32  & Leaky ReLU  \\
        &Convolutional & 32   & Leaky ReLU  \\
        &Convolutional & 16   & Leaky ReLU  \\
        &Convolutional & 16   & Leaky ReLU  \\
        &Convolutional & 1    & None (RD) or Sigmoid (RA) \\ \hline
    \end{tabular}
    \caption{Neural network architectures of the various models. All convolutional layers have a 3x3 kernel. The convolutional layers in the Likelihood model have a stride and padding of 1 to ensure that they preserve the input shape. Upsampling is done by nearest neighbour interpolation. The encoder output represents a 32-dimensional isotropic Gaussian with 32 means and 32 standard deviations.}
    \label{tab:RAArchitectures}
\end{table}

\subsection{Range-azimuth VAE}
Similarly, we extract the RA-map from the raw ADC by using FFT along the fast-time after which we use the Capon method for beam-forming~\cite{Capon} and treat them as inputs and targets of the RA-model (Fig.\ref{fig:model high level}b). Thanks to the similar shape of the input data, the RA-model can share the same architecture as the RD-model.

\subsection{Chirp VAE}
The third model we propose is one that takes the first two raw radar chirps and encodes them in a latent space as in B and C. In this case however, two different decoders are used to decode the latent value either into a RA-map or an RD-map (Fig. \ref{fig:model high level}c). This way, the model can learn by itself the relevant features necessary to create these maps instead of depending upon the predefined Fourier features. This model reuses the same decoder architecture as both the RA and RD models, however it needs a different encoder. The parameters of which can be found in Table \ref{tab:RAArchitectures}. On the same architecture we also experimented with transforming the raw ADC samples through the Range-FFT transform, we call this model Chirp VAE + FFT further in the discussion, and then taking the amplitude and phase values of the resulting signal.

\subsection{Training}
We train all models using the Adam optimizer~\cite{kingma2017adam} with initial learning rate of 1e-4 and use the GECO~\cite{rezende2018taming} optimization technique which puts more weight on the reconstruction term at the start of training to avoid posterior collapse. As a pre-processing step, the range-Doppler and range-azimuth maps are rescaled and squashed using a Sigmoid function to ensure inputs between 0 and 1 as neural network input. We train the models for 10k epochs using the scenario 1 dataset only.

\section{Experimental results}
\label{experiments}
In order to validate the effectiveness of our learned representations, we define four downstream tasks for which we obtained ground truth information. These tasks can be addressed using traditional radar processing techniques and focus on estimating range information (i.e. distance to wall or target), velocity information (i.e. forward moving velocity) and angular information (i.e. angle of arrival w.r.t target). To evaluate the different approaches at test time on the down stream tasks, we fix the weights of the encoder models and in addition train a fully connected neural network with two layers of 128 hidden neurons, in order to regress the task target values. We then report the test set performance and compare to the hand-crafted radar processing baselines. In what follows, we first introduce the different tasks, how we acquired ground truth and performed baseline processing, after which we discuss our results. 

\begin{table*}[t!]
    \centering
    \begin{tabular}{|l||c|c|c|c|c|c|c|c|c|c|c|c|}
        \hline& \multicolumn{3}{|c}{Distance (i)} & \multicolumn{3}{|c}{Velocity (ii)} & \multicolumn{3}{|c}{Target Distance (iii)} & \multicolumn{3}{|c|}{Target Angle (iv)} \\\hline
        model & RMSE & Med. & IntQuart.  & RMSE & Median & IntQuart.  & RMSE & Med. & IntQuart.  & RMSE & Med. & IntQuart.  \\\hline\hline
        radar & 1.32 & 0.43 & 0.51 & 0.20 & -0.02 & 0.28 & 0.48 & 0.20 & 0.14 & 0.14 & 0.03 & 0.07  \\
        RDVAE & \textbf{0.75} & -0.05 & 0.59 & \textbf{0.14} & 0.02 & 0.17 & \textbf{0.19} & 0.03 & 0.22 & 0.17 & 0.02 & 0.19   \\
        RAVAE & 0.86 & 0.00 & 0.70  & 0.28 & -0.06 & 0.37 & 0.23 & 0.04 & 0.25 & \textbf{0.11} & -0.01 & 0.13 \\
        CVAE  & 0.84 & 0.00 & 0.75  & 0.48 & -0.19 & 0.61 & 0.28 & 0.08 & 0.31 & 0.18 & 0.00 & 0.20  \\
        CVAE + FFT & 0.89  & 0.00  & 0.87 & 0.48 & -0.16 & 0.57 & 0.24 & 0.06 & 0.32 & 0.18 & 0.01 & 0.20 \\
        \hline
    \end{tabular}
    \caption{Experimental results. We compare the various models on the four downstream tasks in terms of RMSE, Median error between ground truth and prediction as well as the inter-quartile difference of the error.}
    \label{tab:results}
\end{table*}

\subsection{Downstream tasks}
\label{sec:tasks}
\begin{figure}[b!]
    \centering
    \subfloat[]{\includegraphics[width=0.24\textwidth]{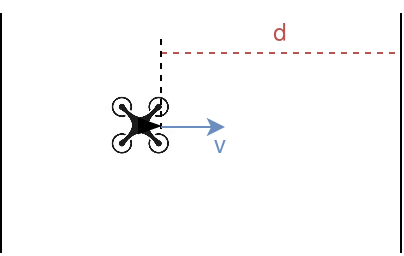}} \hfill
    \subfloat[]{\includegraphics[width=0.24\textwidth]{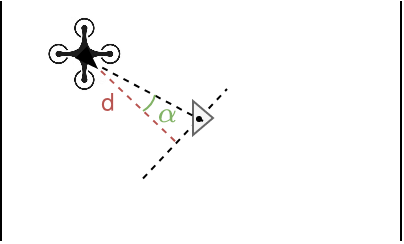}}
    \caption{We consider four downstream tasks: (a) estimate distance $d$ (i) to the wall and forward drone velocity $v$ (ii), and (b) estimate distance $d$ (iii) and angle $\alpha$ to corner reflector (iv).}
    \label{fig:tasks}
\end{figure}

\subsubsection{Distance to wall}

For the first task we use the data collected in scenario 2, in which the drone flies straight ahead towards a wall, makes a U-turn and flies towards the other side. The task is to regress the distance towards the wall, as depicted on Fig.~\ref{fig:tasks}a. To obtain ground truth data, we use pose information of the MOCAP system to calculate the distance to the wall. As radar processing baseline, we detect the highest peak in the range-azimuth map, and calculate the distance based on the range bin and the radar range resolution. 

\subsubsection{Forward velocity}

In addition to estimating the distance to the wall, we also use the data from scenario 2 to estimate the forward velocity of the drone flying towards the wall. The ground truth is established by averaging the distance covered between the previous and next time step, divided by the time interval provided by the timestamps. As traditional radar processing baseline, we now detect the highest peak in the range-Doppler map, and estimate the velocity from the Doppler bin. 

\subsubsection{Distance to corner reflector target}

For target detection we use the data collected in scenario 3, in which a corner reflector was put in the middle of the space, and the drone hovers at various ranges and angles w.r.t. the reflector. As the corner reflector was mounted at the origin of the MOCAP reference frame, the ground truth distance to target is recovered from the drone pose. As traditional radar processing baseline, we track the highest peak in the range-azimuth map (only considering ranges between 0 and 5m).

\subsubsection{Angle to corner reflector target}

\begin{figure}[b!]
        \centering
        \includegraphics[width=0.5\textwidth]{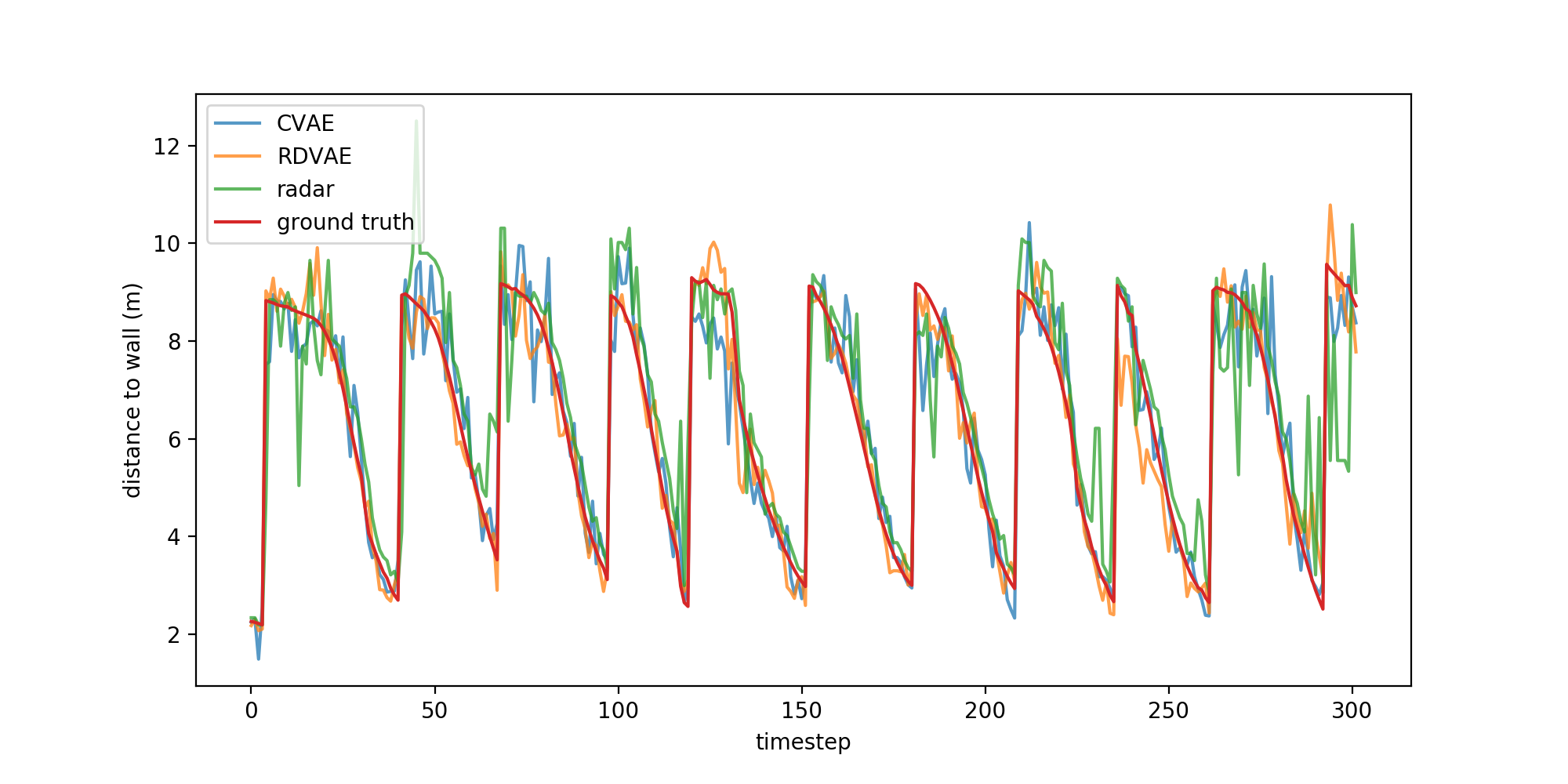}
        \caption{Ground truth distance to wall compared with radar processing, CVAE and RDVAE.}
        \label{fig:distance}
\end{figure}

In addition to the target distance, we also recover the angle of arrival of the corner reflector as shown on Fig.~\ref{fig:tasks}b. In this case, the traditional radar processing baseline tracks the angle bin of maximal amplitude of the reflector peak in the range-azimuth map. 

\subsection{Results}

For all tasks, we adopt a 5/6 to 1/6 train-test split ratio, and train a separate downstream-task neural network for each target task for 500 epochs. We report the test set RMSE against the ground truth, together with the median error as well as the inter-quartile difference. Table~\ref{tab:results} summarizes the results. The best performing model for the distance and velocity tasks is the RDVAE, which is trained on range-Doppler maps. Note that in general the radar processing methods provide a strong baseline, but result in a worse RMSE. This is due to a number of outliers, i.e. when a faulty peak is detected in the spectrum, and is reflected by the lower inter-quartile difference. Also, for the distance estimation task, the radar processing baseline suffers from a bias w.r.t. the ground truth signal, as it overestimates the distance due to the drone not being perfectly perpendicularly aligned to the opposing wall when moving forward. This is reflected in the significant median error. Fig.~\ref{fig:distance} compares the results of the traditional radar processing baseline, the RDVAE and the CVAE against the ground truth. This illustrates how radar processing has a slight overestimate for larger distances, and suffers from some outliers in the distance predictions.

\begin{figure}[htbp]
        \centering
        \includegraphics[width=0.5\textwidth]{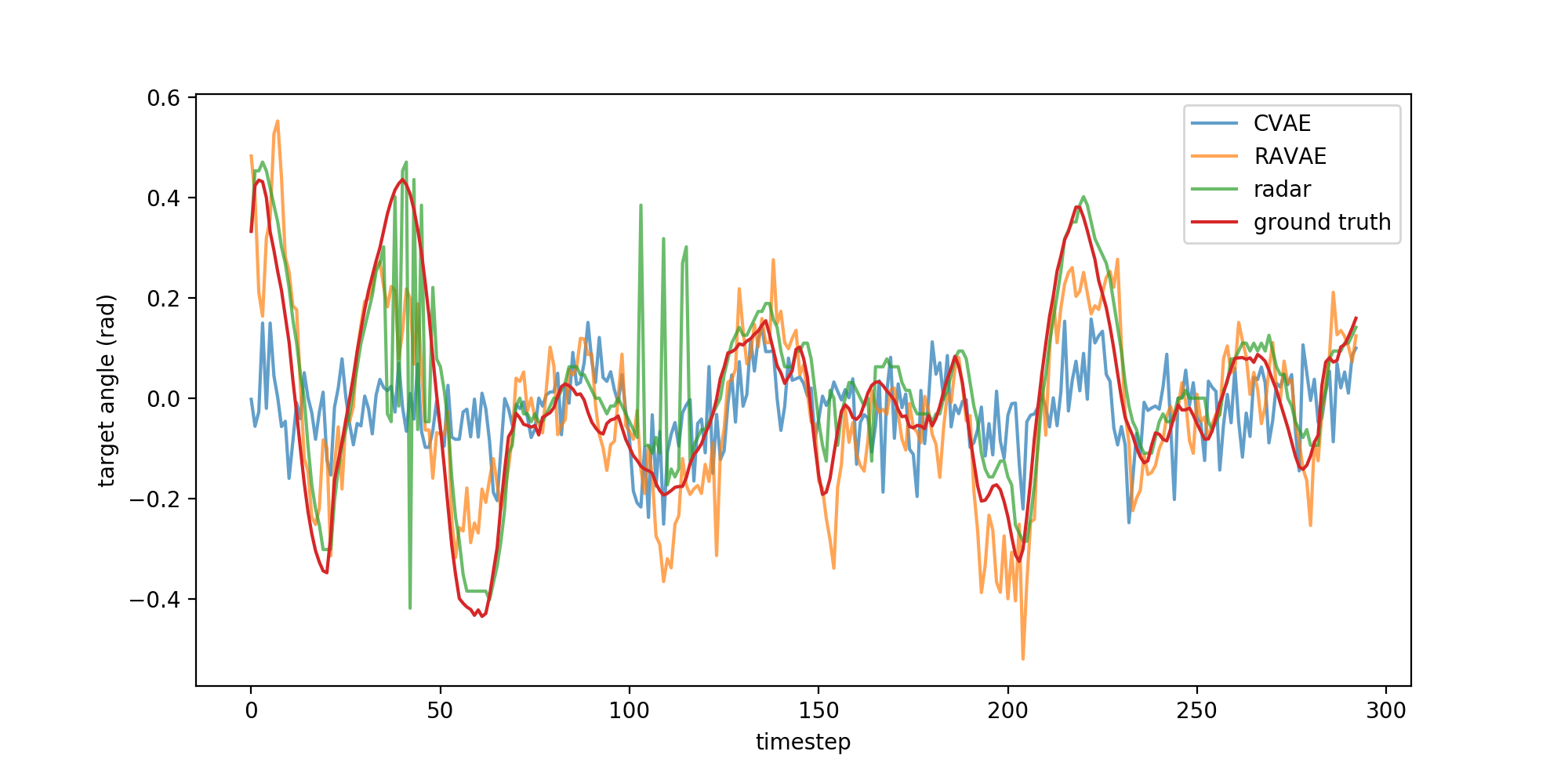}
        \caption{Ground truth angle to target compared against radar processing, CVAE and RAVAE.}
        \label{fig:angle}
\end{figure}

For angular estimation the RAVAE model trained on range-azimuth maps provides the lowest RMSE. Fig.~\ref{fig:angle} compares the ground truth angle over time with CVAE, RAVAE and radar processing. Although radar processing closely matches the ground truth signal most of the time, occasional outliers again yield a large RMSE.

Although the CVAE models perform worse, they do capture some of the characteristics of the distance, velocity and angular information (see Fig. ~\ref{fig:distance} and ~\ref{fig:angle}). Given that the CVAEs only requires the radar to send and receive two chirps and adopts a less complex encoder model, they could provide a lower latency and lower power alternative to full resolution range-azimuth-Doppler maps, at the cost of less accurate predictions.

\subsection{Discussion}

It is important to note that the results of Table \ref{tab:results} are obtained despite training the unsupervised models on the scenario 1 data, which has different characteristics to the data recorded in scenarios 2 and 3. The flights between the aisles contain more clutter and reflections compared to the flights in the open space. This restriction was due to the fact that we required ground truth positioning information for the downstream tasks in order to fairly conduct our evaluation. It is interesting to note that despite this discrepancy, the learned representations still capture the necessary information to address the downstream tasks. An important point of future work is to record data in a more diverse set of environments in order to further evaluate to what extent our learned representations can generalize. In addition, one could think of more complex scenarios and downstream tasks, such as detecting moving obstacles or detecting loop closures in a simultaneous localization and mapping (SLAM) setting.

Despite the inferior performance of the models acting on raw ADC data, these might still be an interesting avenue for future research, given that they need 64 times less chirps and less pre-processing steps. For example, different chirp encoder architectures using e.g., recurrent layers could be investigated for refining the radar representations. Doing so, an adaptive power versus accuracy trade-off could be provided during radar processing, by only increasing the number of chirps to improve the accuracy of the system when needed.

\section{Conclusion}
\label{conc}
In this paper, we have proposed a method to learn low-dimensional representations of FMCW radar data. We have experimented with a number of encoder-decoder architectures respectively operating on range-Doppler maps, range-azimuth maps or raw chirp data. To benchmark the effectiveness of the resulting representations, we have compared neural network regressors against traditional hand-crafted radar processing for a number of downstream tasks. For this purpose, we have presented a novel dataset recording both camera RGBD and FMCW radar data from a drone flying in an indoor warehouse environment. This paper has provided what is, to the best of our knowledge, one the first works that address the issue of learning robust radar representations by developing novel VAE-based neural architectures for processing radar data, and evaluate these on a set of distinct downstream tasks. As future work, we will record data in a more diverse set of environments, in order to also address to what extent such representations can generalize to multiple environments, and experiment with other chirp encoder architectures.


\bibliographystyle{IEEEtran} 
\balance
\bibliography{bibliography} 

\begin{thebibliography}{10}
\providecommand{\url}[1]{#1}
\csname url@samestyle\endcsname
\providecommand{\newblock}{\relax}
\providecommand{\bibinfo}[2]{#2}
\providecommand{\BIBentrySTDinterwordspacing}{\spaceskip=0pt\relax}
\providecommand{\BIBentryALTinterwordstretchfactor}{4}
\providecommand{\BIBentryALTinterwordspacing}{\spaceskip=\fontdimen2\font plus
\BIBentryALTinterwordstretchfactor\fontdimen3\font minus
  \fontdimen4\font\relax}
\providecommand{\BIBforeignlanguage}[2]{{%
\expandafter\ifx\csname l@#1\endcsname\relax
\typeout{** WARNING: IEEEtran.bst: No hyphenation pattern has been}%
\typeout{** loaded for the language `#1'. Using the pattern for}%
\typeout{** the default language instead.}%
\else
\language=\csname l@#1\endcsname
\fi
#2}}
\providecommand{\BIBdecl}{\relax}
\BIBdecl

\bibitem{Indoor}
G.~De~Croon and C.~De~Wagter, ``Challenges of autonomous flight in indoor
  environments,'' in \emph{2018 IEEE/RSJ International Conference on
  Intelligent Robots and Systems (IROS)}, 2018, pp. 1003--1009.

\bibitem{richards2005fundamentals}
M.~A. Richards, \emph{Fundamentals of radar signal processing}.\hskip 1em plus
  0.5em minus 0.4em\relax New York: McGraw-Hill, 2005.

\bibitem{Safa21}
A.~Safa, T.~Verbelen, L.~Keuninckx, I.~Ocket, M.~Hartmann, A.~Bourdoux,
  F.~Catthoor, and G.~G.~E. Gielen, ``A low-complexity radar detector
  outperforming os-cfar for indoor drone obstacle avoidance,'' \emph{IEEE
  Journal of Selected Topics in Applied Earth Observations and Remote Sensing},
  vol.~14, pp. 9162--9175, 2021.

\bibitem{Wheeler2017}
\BIBentryALTinterwordspacing
T.~A. Wheeler, M.~Holder, H.~Winner, and M.~Kochenderfer, ``Deep stochastic
  radar models,'' 1 2017. [Online]. Available:
  \url{http://arxiv.org/abs/1701.09180}
\BIBentrySTDinterwordspacing

\bibitem{Major2019}
B.~Major, D.~Fontijne, A.~Ansari, R.~T. Sukhavasi, R.~Gowaikar, M.~Hamilton,
  S.~Lee, S.~Grechnik, and S.~Subramanian, ``Vehicle detection with automotive
  radar using deep learning on range-azimuth-doppler tensors,'' 2019.

\bibitem{Prahbat21}
P.~K. Rai, H.~Idsøe, R.~R. Yakkati, A.~Kumar, M.~Z. Ali~Khan, P.~K.
  Yalavarthy, and L.~R. Cenkeramaddi, ``Localization and activity
  classification of unmanned aerial vehicle using mmwave fmcw radars,''
  \emph{IEEE Sensors Journal}, vol.~21, no.~14, pp. 16\,043--16\,053, 2021.

\bibitem{Geng2021}
Z.~Geng, H.~Yan, J.~Zhang, and D.~Zhu, ``Deep-learning for radar: A survey,''
  \emph{IEEE Access}, vol.~9, pp. 141\,800--141\,818, 2021.

\bibitem{Patel2019}
K.~Patel, K.~Rambach, T.~Visentin, D.~Rusev, M.~Pfeiffer, and B.~Yang, ``Deep
  learning-based object classification on automotive radar spectra,'' in
  \emph{2019 IEEE Radar Conference (RadarConf)}, 2019, pp. 1--6.

\bibitem{Mostafa2018}
M.~Mostafa, S.~Zahran, A.~Moussa, N.~El-Sheimy, and A.~Sesay, ``Radar and
  visual odometry integrated system aided navigation for uavs in gnss denied
  environment,'' \emph{Sensors}, vol.~18, no.~9, 2018.

\bibitem{Capon}
J.~Li, P.~Stoica, and Z.~Wang, ``On robust capon beamforming and diagonal
  loading,'' \emph{IEEE Transactions on Signal Processing}, vol.~51, no.~7, pp.
  1702--1715, 2003.

\bibitem{Kingma2014}
D.~P. Kingma and M.~Welling, ``Auto-encoding variational bayes,'' \emph{arXiv
  preprint}, 2014.

\bibitem{Rezende2014}
D.~J. Rezende, S.~Mohamed, and D.~Wierstra, ``Stochastic backpropagation and
  approximate inference in deep generative models,'' in \emph{Proceedings of
  the 31st International Conference on Machine Learning}, vol.~32, no.~2.\hskip
  1em plus 0.5em minus 0.4em\relax PMLR, 22--24 Jun 2014, pp. 1278--1286.

\bibitem{kingma2017adam}
D.~P. Kingma and J.~Ba, ``Adam: A method for stochastic optimization,'' 2017.

\bibitem{rezende2018taming}
D.~J. Rezende and F.~Viola, ``Taming vaes,'' 2018.

\end{thebibliography}

\end{document}